\documentclass[lettersize,journal]{IEEEtran}
\usepackage{amsmath,amsfonts}
\usepackage{algorithmic}
\usepackage{algorithm}
\usepackage{array}
\usepackage[caption=false,font=normalsize,labelfont=sf,textfont=sf]{subfig}
\usepackage{textcomp}
\usepackage{stfloats}
\usepackage{url}
\usepackage{verbatim}
\usepackage{graphicx}
\usepackage{cite}
\usepackage{hyperref}
\usepackage{amsthm}
\usepackage{amsopn}

\usepackage{lipsum}
\usepackage{epstopdf}
\usepackage{booktabs}%
\usepackage{subfig}
\usepackage{amssymb,braket,xcolor,upgreek,bbold}
\usepackage{soul}
\ifpdf
  \DeclareGraphicsExtensions{.eps,.pdf,.png,.jpg}
\else
  \DeclareGraphicsExtensions{.eps}
\fi

\newtheorem{corollary}{Corollary}
\newtheorem{proposition}{Proposition}

\newcommand{\be}{\begin{equation}}
\newcommand{\ee}{\end{equation}}

\newcommand{\bJ}{{\bf J}}
\newcommand{\bD}{{\bf D}}

\newcommand{\bI}{{\bf I}}

\newcommand{\bC}{{\bf C}}

\newcommand{\adots}{\mathinner{\mkern2mu\raisebox{1pt}{.}\mkern2mu
  \raisebox{4pt}{.}\mkern2mu\raisebox{7pt}{.}\mkern1mu}}

\begin{document}



\title{A Low-complexity Structured Neural Network to Realize States of Dynamical Systems}

\author{Hansaka Aluvihare\thanks{H. Aluvihare is with the Department of Mathematics, Embry-Riddle Aeronautical University, Daytona Beach, FL, USA. 
Email: aluvihah@my.erau.edu}, Levi Lingsch\thanks{L. Lingsch is with the Seminar for Applied Mathematics, ETH Zurich, Zurich, Switzerland. Email: levi.lingsch@sam.math.ethz.ch.}, Xianqi Li\thanks{X. Li is with the Department
of Mathematics \& Systems Engineering, Florida Institute of Technology, Melbourne, FL, USA. Email: xli@fit.edu}, and Sirani M. Perera \thanks{S. M. Perera is with the Department of Mathematics, Embry-Riddle Aeronautical University, Daytona Beach, FL, USA. Email: pereras2@erau.edu}
\thanks{This work was funded by the Division of Mathematical Sciences, National Science Foundation with the award numbers 2410676 \& 2410678.}
\thanks{Manuscript received July, 2025.}}



\maketitle

\begin{abstract}
Data-driven learning is rapidly evolving and places a new perspective on realizing state-space dynamical systems. 
However, dynamical systems derived from nonlinear ordinary differential equations (ODEs) suffer from limitations in computational efficiency. 
To address this, we propose a structured neural network (StNN) that uses structured matrix theory and relies on a Hankel operator derived from time-delay measurements to solve dynamical systems. Specifically, the StNN identifies an optimal representation using the Hankel operator, providing a more computationally efficient alternative to existing data-driven approaches. We show that the proposed StNN places an optimal solution answering the demand for inference time, number of parameters, and computational complexity compared with the conventional neural networks and also with the classical data-driven techniques, such as Dynamic Mode Decomposition(DMD), Sparse Identification of Nonlinear Dynamics (SINDy), and Hankel Alternative View of Koopman (HAVOK), which is commonly known as delay-DMD or Hankel-DMD. 
Furthermore, we present numerical simulations to solve dynamical systems utilizing the StNN based on structured matrix theory followed by the Hankel operator beginning from the fundamental Lotka-Volterra model, where we compare the StNN with the LEarning Across Dynamical Systems (LEADS), and extend our analysis to highly nonlinear and chaotic Lorenz systems, comparing the StNN with conventional neural networks, DMD, SINDy, and HAVOK.
Hence, we show that the proposed StNN paves the way for realizing state-space dynamical systems with a low-complexity learning algorithm, enabling prediction and understanding of future states.
\end{abstract}



\section{Introduction}
\label{intro}
\IEEEPARstart{M}{athematical} 
models can be utilized to continually analyze the dynamics of system states, providing a unique tool to represent dynamical systems. These models are formulated through a set of rules, often expressed as differential or difference equations, which dictate how the state variables evolve through time in continuous or discrete settings. Describing the evolution of state variables over time is a key aspect of solving dynamical systems. Depending on the system's complexity and continuity, one could achieve this while analytically solving the systems. Various methods can be used to analyze the solutions for continuous dynamical systems. These include separating variables for simple systems, linearizing to approximate nonlinear systems, spectral analysis, employing phase plane, and utilizing Laplace or Fourier transformations to obtain efficient solutions \cite{GoVa13,K16,TB97,D97}. These techniques provide a comprehensive understanding of the behavior and characteristics of dynamical systems. On the other hand, numerical techniques such as Euler's method, Runge-Kutta method, finite difference, finite element method, and spectral analysis are well-known to be applied to solve dynamical systems using iterative formulas \cite{Ascher1998,Thomas1995}.


%

The exponential growth in data science places a new perspective on realizing state-space dynamical systems through data-driven approaches \cite{SBNK19,St19}. With this said, dynamic mode decomposition (DMD) and extended DMD are utilized to identify the spatiotemporal structure of the high-dimensional data incorporating SVD through dimension reduction \cite{KBBP16}. The DMD offers a modal decomposition, in which each mode is composed of spatially correlated structures that exhibit identical linear behavior over time. Thus, DMD not only reduces dimensions by using a smaller set of modes but also provides a model for the evolution of these modes over time and can be utilized to obtain best-fit linear models \cite{Schmid2008DynamicMD,schmid_2010,JonathanHTu2014}. Identifying the nonlinear structure and parameters of dynamical models from data can be expensive due to the combinatorial possibilities for analyzing structures. Fortunately, the Sparse Identification of Nonlinear Dynamics (SINDy) algorithm provides a way to bypass costly searches by exploring the dependence of functional variables in the system \cite{Brunton2015DiscoveringGE}. On the other hand, Koopman operator theory presents an alternative perspective of dynamical systems in terms of the evolution of measurements because it is possible to represent a nonlinear dynamical system in terms of an infinite-dimensional linear operator acting on a Hilbert space of measurement functions of the state of the system \cite{Chen2012VariantsOD,Brunton2021ModernKT}. 

Machine learning (ML) and deep learning (DL) algorithms have emerged as powerful tools for modeling, predicting, and controlling dynamical systems, offering significant advantages over classical methods in handling nonlinearity, high dimensionality, and uncertainty. Recent advances in neural networks, Convolutional Neural Networks (CNN), and Recurrent Neural Networks (RNN) have demonstrated outstanding success in capturing temporal dependencies and chaotic behaviors in dynamical systems \cite{brunton2022data,rajendra2020modeling}. Lusch et al. \cite{lusch2018deep} used an autoencoder-based deep learning framework to discover Koopman eigenfunctions from data, enabling globally linear representations of nonlinear dynamics on low-dimensional manifolds. Moreover, \cite{atencia2005hopfield} presented a Hopfield neural network-based method for online parameter estimation in system identification, featuring time-varying weights and biases to handle dynamic target functions. The simulations demonstrate better performance over classical gradient methods, achieving lower errors. A convolutional autoencoder and a multi-timescale recurrent neural network-based method are proposed in \cite{suzuki2018motion} for flexible behavior combination in robots using dynamical systems based on point attractors, incorporating instruction signals and phases to divide tasks into subtasks. Moreover, \cite{trischler2016synthesis} proposed a feedforward network on a dynamical system's vector field using backpropagation, then converted it into a continuous-time RNN, demonstrating its effectiveness through numerical examples. Physics-informed neural networks (PINNs) have gained significant attention in recent years, \cite{antonelo2024physics} introduced Physics-Informed Neural Nets for Control, a novel framework extending traditional PINNs by incorporating initial conditions and control signals. It utilizes an autoregressive self-feedback method to provide accurate and adaptable simulations, as proven on nonlinear systems such as the Van der Pol oscillator with faster inference.

This paper presents a low-complexity structured neural network (StNN) designed to learn the dynamics of state-space systems and predict future states using a Hankel operator derived from a time-delay series of state measurements. We emphasize that Hankel matrices possess a unique structure, which can be effectively leveraged to solve systems of linear equations using low-complexity algorithms \cite{HR84,P01,BS16,O03,GH01,BLV98}. Hankel matrices have been utilized in spectrum analysis, spectral decomposition, the evaluation of linear and chaotic stochastic dynamics, and the realization of state space systems \cite{K16,Broomhead,Juang1985AnER,Mezi2005SpectralPO}. Furthermore, the modern Koopman operator theory presents a compelling approach by employing delay embedding-based Hankel matrices as accurate computational tools for modeling dynamical systems, which is commonly known as delay-DMD or Hankel-DMD \cite{Arbabi2016ErgodicTD,Brunton2014ExtractingSC,JonathanHTu2014}. 

The paper is organized as follows. We propose a simple structured operator called the Hankel operator and utilize it to solve non-linear systems of ODEs using efficient computations in

\begin{figure*}[ht]
    \centering
    \includegraphics[width=1.0\linewidth]{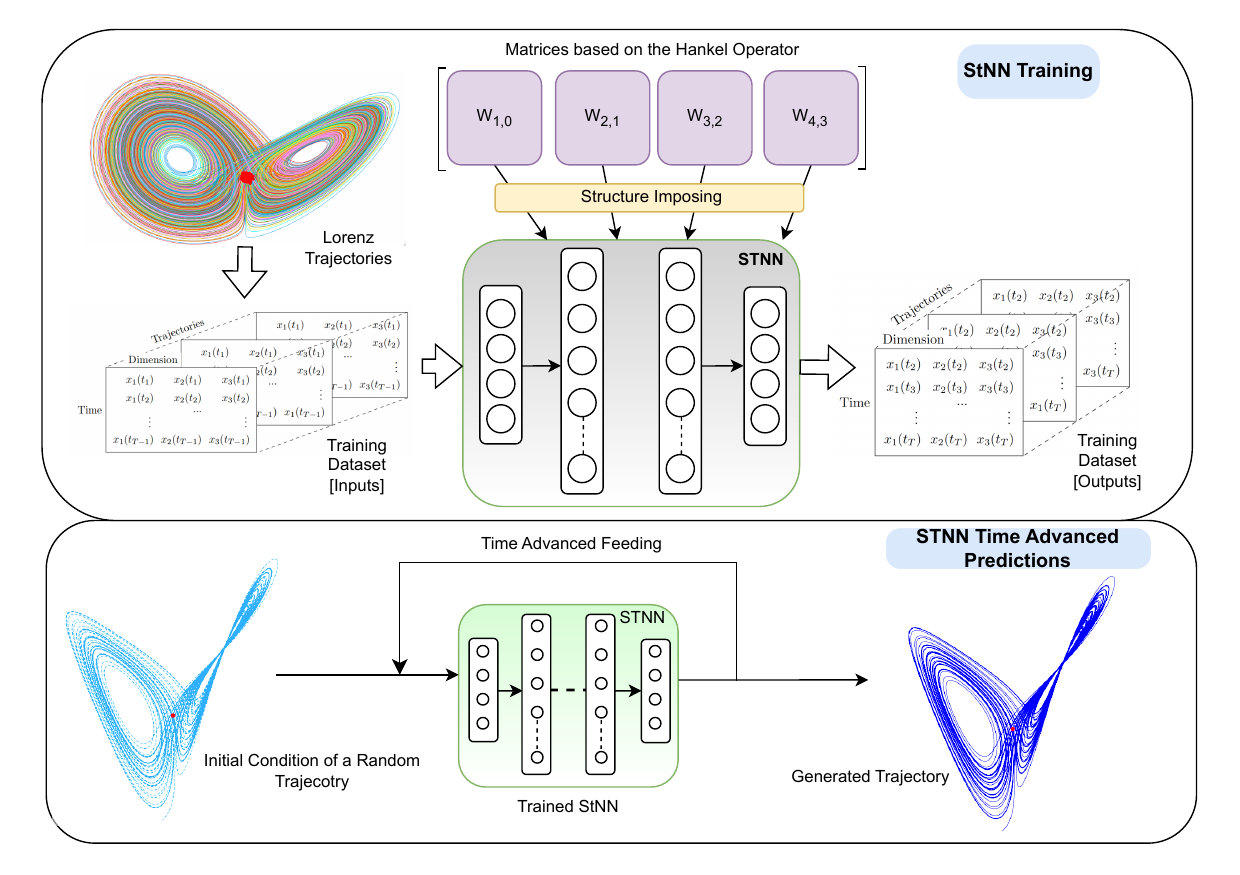} 
    \caption{An overview of the Structured Neural Network (StNN) framework for modeling dynamical systems. The top part depicts the StNN's training process, which uses Lorenz system trajectories to create a dataset and structure-imposing matrices based on the Hankel operation to guide the learning process. The bottom portion shows time-advanced predictions, in which a trained StNN creates future trajectories based on an initial condition of a random trajectory}
    \label{fig:overview}
\end{figure*}

\section{Preliminaries \& Factorizations: Learning and Realizing States for Dynamical Systems}
\label{sec:main}
We propose to explore the chaotic behavior of dynamical systems by learning a real-valued Hankel structured operator. We note here that the structure-imposed operator is the key to proposing low-complexity learning. Thus, we will utilize data-driven learning to understand dynamical systems while proposing a low-complexity neural network called a StNN. Let us start the section by introducing notations which we will utilize frequently in the paper.

\subsection{Frequently Used Notations}
\label{sub:freqn}
Here we introduce notations for sparse and orthogonal matrices which will frequently be used in this paper. We first define states of dynamical systems at time $t_k$ by
\begin{equation}
{\bf x}_k=\begin{bmatrix}
    x_1(t_k) & x_2(t_k) & \cdots & x_n(t_k)
\end{bmatrix}^T, 
    \label{state}
\end{equation}
where $^T$ for the transpose, and $k=0, 1, \cdots, n-1$. We utilize time-delays series of a state measurement $\{x(\tau_{k})\}_{k=0}^{n-1}$ 
to define 
a Hankel operator ${\bf H} \in \mathbb{R}^{n \times n}$ s.t. 
\begin{equation}
{\bf H}:=\begin{bmatrix}
x(\tau_0) & x(\tau_1) & x(\tau_2) & \cdots & x(\tau_{n-2}) & x(\tau_{n-1}) \\
x(\tau_1) & x(\tau_2) & x(\tau_3) & \cdots & x(\tau_{n-1}) & x(\tau_{n-2}) \\
x(\tau_2) & x(\tau_3) & x(\tau_4) & \cdots & x(\tau_{n-2}) & x(\tau_{n-3}) \\
\vdots & \vdots & \vdots & \vdots & \adots & \vdots \\
x(\tau_{n-2}) & x(\tau_{n-3}) & x(\tau_{n-4}) & \cdots  & x(\tau_{2}) & x(\tau_{1}) \\
x(\tau_{n-1}) & x(\tau_{n-2}) & x(\tau_{n-3}) & \cdots  & x(\tau_{1}) & x(\tau_{0}) \\
\end{bmatrix},
    \label{mformope}
\end{equation}
where $\tau_k$'s are time-delay measurements. 
We note here that the matrix ${\bf H}$ (\ref{mformope}) differs from HAVOK \cite{SBNK19}. In HAVOK, the elements of Hankel matrix are defined based on an evaluation of states using the Koopman operator $\kappa$, specifically,  the first column and row are defined as $[x(t_1), \kappa x(t_1), \cdots, \kappa^{q-1} x(t_1)]$ and $[x(t_1), \kappa x(t_1), \cdots, \kappa^{p-1} x(t_1)]^T$, respectively, where $p$ and $q$ are constants.

We also define the DFT matrix by
${\bf \mathfrak{F}}_n = \frac{1}{\sqrt{n}}\:[ w_n^{kl}]_{k,l=0}^{n-1}$,
where $w_n=e^{-\frac{2\pi i}{n}}$ is the primitive $n^{\rm th}$ root of unity, 
a scaled DFT matrix by $\tilde{\bf \mathfrak{F}}_n=\sqrt{n} \:{\bf \mathfrak{F}}_n$ and its conjugate transpose by ${\bf \mathfrak{F}}^*_n$, 
a highly sparse matrix by ${\bJ}_{r \times n} = \left [\begin{array}{c}
\bI_n \\ 
\hline
{\bf 0}_n
\end{array}  \right ]$ where $r=2n$, $\bI_{n}$ is the identity matrix and ${\bf 0}_{n}$ is the zero matrix, an antidiagonal matrix by $\tilde{\bI}_{n}$,
a diagonal matrix by 
$
\breve{\bD}_{r}={\rm diag}\left[\tilde{\bf \mathfrak{F}}_{r}{\bf c} \right]
$
where a circulant matrix $\bC_{r}$ defined by the first column ${\bf c}$ s.t.
$
\begin{aligned}
{\bf c}  =[x(\tau_{n-1}), x(\tau_{n-2}),  \cdots, x(\tau_{0}), x(\tau_{n-1}), \\
x(\tau_{0}),  x(\tau_{1}), x(\tau_{2}), \cdots, x(\tau_{n-2}) ]^T.
\end{aligned}
$

\subsection{Preliminaries: Dynamical Systems and Operator}
\label{sub:prelim}
This section introduces fundamentals related to dynamical systems derived from nonlinear ordinary differential equations (ODEs). We will also discuss an operator designed to effectively solve these dynamical systems. 
One could say that the nonlinear ODEs represent the dynamical system of the form 
\begin{equation*}
    \frac{d}{dt} {\bf x}(t)={\bf f} ({\bf x}(t), t),
    \label{dyeq}
\end{equation*}
where ${\bf x}(t) \in \mathbb{R}^n$ is the state of the system evolving in time $t$ and ${\bf f}$ is a vector-valued function. As in the systems of linear equations, one could also answer the question of the existence and uniqueness of the dynamical systems. In this situation, this could be generally achieved by analyzing the Lipschitz continuity of the function ${\bf f}$. On the other hand, the discrete-time dynamical systems are of the form 
\begin{equation*}
    {\bf x}_{k+1}={\bf F} ({\bf x}_{k}),
    \label{dtds}
\end{equation*}
and it sees the states of the system at the $k^{\rm th}$ iteration as ${\bf x}_{k} \in \mathbb{R}^n$ having a non-linear function ${\bf F}$, which will usually denote iterations forward in time, so that ${\bf x}_{k}={\bf x} (k \Delta t)$. This is the situation, in which we could seek the solution of a dynamical system as a solution of a system of linear equations. Thus, to sum up, many problems in dynamical systems ultimately lead to a solution of systems of linear equations.     
On the other hand and due to the nonlinear nature of these dynamical systems, we propose a learning algorithm to train a neural network so that the network could learn an updated state from ${\bf x}_k$ (\ref{state}) to ${\bf x}_{k+1}$ 
using a low-complexity ML operator. 

To learn an operator ${\bf H}$, we start with the discrete-time dynamical system ${\bf x}_{k+1}={\bf F} ({\bf x}_{k})$ while defining the function evaluation as the matrix-vector computation via ${\bf F}: \mathbb{R}^n \rightarrow \mathbb{R}^n$ s.t. ${\bf F}({\bf x})={\bf H}{\bf x}$. Thus, the states of the system as it evolves in time can be defined via
\begin{equation}
{\bf x}_{k+1}={\bf F} ({\bf x}_{k})={\bf H} {\bf x}_{k},
    \label{dtol}
\end{equation} 
We note that the Hankel operator is linear with respect to states $\bf x$ and $\tilde{\bf x}$  s.t. ${\bf H}(c_1 {\bf x}+c_2 \tilde{\bf x})=c_1{\bf H}{\bf x} + c_2{\bf H}\tilde{\bf x}$, where $c_1$ and $c_2$ are constants. Thus, to obtain the best-fit Hankel operator that best advances snapshot state measurements forward in time, we propose a best-fit operator as defined in the next section.

Thus, having a rich set of information based on the time-delayed operator ${\bf H}$ to predict future states of chaotic systems leads to better prediction than linear or nonlinear systems with trajectories trapped at fixed points or on periodic orbits \cite{SBNK19}. On the other hand, instead of advancing linear or non-linear measurements of the states of a system, like in the DMD, we could measure time-delayed measurements using the Hankel operator ${\bf H}$ following the HAVOK \cite{Arbabi2016ErgodicTD,Brunton2016ChaosAA} and utilize that to obtain low-complexity algorithms to realize state measurements as in next section. 

\subsection{Learn a Best-fit Operator}
We propose to obtain a best-fit operator for ${\bf H}$--say $\widehat{\bf H}$, determined via time snapshots of spatiotemporal data. 
Furthermore, we propose to enhance learning by capturing the evolution of the nonlinear dynamical system using data-driven embedding based on the best-fit operator.

Let us obtain the best-fit operator  $\widehat{\bf H}$ determined via time-delays series of a state measurement $\{x(\tau_{k})\}_{k=0}^{n-1}$ to optimize the data-driven learning. 

\begin{proposition}
\label{thir:prop}
Let ${\bf {X}}_{l,k}=[{x}_{l}(\tau_k)]_{l=1,k=0}^{n,n-1}$ is the time-delay snapshots matrix, ${\bf {X'}}_{l,k}=[{x}_l({t}_k)]_{l=1,k=0}^{n,n-1}$ is the time-advanced snapshots matrix, ${t}_{k}=\tau_k+\Delta t$, and $\Delta t$ is the timestep. Then, an approximate  solution for the Hankel operator ${\bf H}$--say $\widehat{\bf H}$ can be obtained via
\begin{equation}
{\bf {X'}} \approx \widehat{\bf H} {\bf X}, \quad \widehat{\bf H}={\rm argmin}_{\bf H}
\bigl\{ 
 \frac{1}{2}||{\bf {X'}} - {\bf H}{\bf X}||^2_F + \alpha||{\bf H}||_\eta \bigr\} ,
\label{meq}
\end{equation}
where $\| \cdot \|_F$ is the Frobenius norm, $\|\cdot\|_\eta$ represents the nuclear norm for low-rank matrices, and $\alpha$ is a non-negative tuning parameter controlling the regularization of the low-rank matrix.
\end{proposition}
\begin{proof}
Without loss of generality, we consider ${\bf H}^T$: the transpose of the Hankel operator ${\bf H}$ since the singular values of the ${\bf H}^T$ are equal to those of  ${\bf H}$. Now \eqref{meq} is equivalent to the following formulation
\begin{equation}
\begin{split}
&{(\bf {X'})^T} \approx {\bf X}^T\widehat{{\bf H}^T } \: \quad \\&{\rm where}, \\
&\: {\widehat{{\bf H}^T}}={\rm argmin}_{{\bf H}^T}
\bigl\{ 
 \frac{1}{2}||({\bf {X'}})^T - {\bf X}^T{\bf H}^T||^2_F + \alpha||\bf {\bf H}^T||_\eta \bigr\} ,
\label{meq1}
\end{split}
\end{equation}
which is a convex optimization problem due to the fact that the nuclear norm $\|\cdot\|_\eta$ is a convex relaxation of the rank minimization problem \cite{toh2010accelerated}. Moreover, because this norm is coercive, there exists an optimal solution for \eqref{meq1}. By the framework in \cite{ji2009accelerated}, the formulated optimization problem \eqref{meq1} can be solved equivalently as 
\begin{equation}
\begin{split}
 &{\widehat{{\bf H}^T}}= {\rm argmin}_{{\bf H}^T}
\bigl\{ 
 ||{\bf H}^T-  \\ & \left({\bf H}^T_{k-1}- \frac{1}{t_k}{\bf X}({\bf X}^T{\bf H}^T_{k-1}-({\bf {X'}})^T)\right)||^2_F +\frac{2\alpha}{t_k}||\bf {\bf H}^T||_\eta \bigr\} ,
\label{meq2}
\end{split}
\end{equation}
where ${\bf H}^T_{k-1}$ is the $k-1$ iterates for $H$ and $t_k$ is the stepsize. The minimization problem \eqref{meq2} can be solved by computing the singular value decomposition (SVD) of $\left({\bf H}^T_{k-1}-\frac{1}{t_k}{\bf X}({\bf X}^T{\bf H}^T_{k-1}-({\bf {X'}})^T)\right)$. Then the soft-thresholding operator can be applied on the singular values. By Theorem 2.1 in \cite{cai2010singular}, the approximate solution for \eqref{meq2} has low-rank properties, which can be chosen as an approximate solution for the Hankel operator.
\end{proof}

Once the data-driven dynamical system has evolved, it is possible to further enhance the algorithms to differentiate between the inherent, spontaneous dynamics and the impact of actuation. This differentiation amounts to a more comprehensive evolution equation \cite{Proctor2014DynamicMD}
\begin{equation}
 {\bf x}_{k+1} \approx \widehat{\bf H} {\bf x}_k+{\bf G} {\bf u}_k,  
 \label{bbeq}
\end{equation}
where $\widehat{\bf H}$ is an $n \times n$ system matrix realized as the best-fit Hankel operator,  ${\bf G}$ is an ${n \times q}$ input matrix and ${\bf u}_k=\begin{bmatrix}
    u_1(t_k) & u_2(t_k) & \cdots & u_q(t_q)
\end{bmatrix}^T \in \mathbb{R}^q$ is an input vector. The system extension (\ref{bbeq}) stems from (\ref{meq}) leads to seek time-advanced states using time-delayed states-based Hankel operator.

\subsection{Factorize the Hankel Operator to Realize State Measurements}
\label{sub:fp}
In this section, we propose to utilize the factorization of the Hankel operator to realize state measurements using low-complexity algorithms.
This is due to the fact that the data-driven approaches are computationally intensive, despite the potential for low-rank approximation via established SVD techniques. However, Hankel is a structured matrix, which allows us to explore an alternative approach for low-rank approximation in HAVOK. Instead of depending on SVD, we propose utilizing low-complexity algorithms that leverage the inherent structure of the Hankel operator to observe time-advanced states. This approach aims to reduce the complexity of training data-driven models efficiently.

\begin{proposition}
 \label{first:propo}
Let ${\bf H}$ be the Hankel operator (\ref{mformope}) determined via  time-delays series of a state measurement $\{x(\tau_{k})\}_{k=0}^{n-1}$. Then, the Hankel operator can be calculated through the following low-rank matrices
 \begin{equation}
 \begin{aligned}
{\bf H} = {\bf H}_l + {\bf H}_u-x(\tau_{n-1}) \tilde{\bf I}_n,
\\ \nonumber
{\bf H}_u = [\tilde{\bf x}, Z\tilde{\bf x}, \cdots, Z^{n-1}\tilde{\bf x}], \:\: {\rm and}\:\: {\bf H}_l=\tilde{\bf I}_n [{\bf H}_u]^T\tilde{\bf I}_n,
\end{aligned}
     \label{sumeqnr1}
 \end{equation}
 where ${\bf Z}$ is $n \times n$ upper shift matrix and $\tilde{\bf x}=[x(\tau_0), x(\tau_1), x(\tau_2), \cdots, x(\tau_{n-1})]^T$  
\end{proposition}
\begin{proof}
The operator ${\bf H}$ is a per-symmetric Hankel matrix determined by the first column(or row) of ${\bf H}$ s.t. $[x(\tau_0), x(\tau_1), \cdots, x(\tau_{n-1})]^T$, and when $Z$ is the lower shift matrix and when $\tilde{\bf x}_{n \times 1}$ is defined as above, we could write ${\bf H}_u=[\tilde{\bf x}, Z\tilde{\bf x}, \cdots, Z^{n-1}\tilde{\bf x}]$ followed by the per-transpose to get ${\bf H}_l$ with the addition of $x(\tau_{n-1}) \tilde{\bf I}_n$ to obtain ${\bf H}$. 
\end{proof}

\begin{corollary}
 \label{first:complex}   
 Let the Hankel operator ${\bf H}$ be utilized to advance a snapshot of states forward in time
 using Propositions \ref{thir:prop} and \ref{first:propo}, then complexity 
 in realizing time-advanced states cost $\mathcal{O}(n^s)$, where $1<s<2$. 
\end{corollary}

\begin{proof}
Since ${\bf H}$ is the structured matrix determined by $\mathcal{O}(n)$ elements, we could compute ${\bf H} {\bf x}_k$ by utilizing the upper shift matrix ${\bf Z}_{r \times r}$ followed by the vector $\tilde{\bf x}_{r \times 1}$ in (\ref{sumeqnr1}) to reduce the complexity in computing the conventional matrix-vector product of  ${\bf H}{\bf x}_k$ from $\mathcal{O}(n^2)$ to $\mathcal{O}(n^s)$, where $1<s<2$.      
\end{proof}

By utilizing the radix-2 algorithm to compute the Toeplitz matrices by a vector using 2-FFTs \cite{SLARSN21,B19} as opposed to 3-FFTs \cite{KS99,SLAL22} for an even length s.t. $n=2^p (p \geq 1)$, and also computing the odd order Toeplitz matrices by a vector using 2-FFTs in \cite{PERERA2025}, we could also state the following factorization to decompose the Hankel operator ${\bf H}$ using 2-FFTs. 

\begin{proposition}
\label{secon:propo}
Let ${\bf H}$ be the Hankel operator (\ref{mformope}) determined via 
 time-delays series of a state measurement $\{x(\tau_{k})\}_{k=0}^{n-1}$. Then, the operator can be realized using the following decomposition
\begin{equation}
{\bf H}=
\tilde{\bf I}_{n}  [{\bf J}^T]_{n \times r} {\bf C}_{r} [{\bf J}]_{r \times n},
\label{MLFFT}
\end{equation}
where the circulant matrix ${\bf C}_{r}={\bf \mathfrak{F}}^*_r \breve{\bD}_{r} {\bf \mathfrak{F}}_r$.
\end{proposition}
\begin{proof}
When $n=2^p(p \geq 1)$, we could compute ${\bf H}$ by using the 2-FFTs as described in \cite{SLARSN21}, and when $n \neq 2^p$ we could pad with zeros to the nearest but the greatest power of 2 followed by the use of 2-FFTs in \cite{SLARSN21}. 
\end{proof}

We note here that one could compute odd-length ${\bf H}$ using 2-FFTs as described in \cite{PERERA2025}.

\begin{corollary}
 \label{second:complex}   
 Let the Hankel operator ${\bf H}$ be utilized to advance the observation of states from ${\bf x}_k$ to ${\bf x}_{k+1}$ using Propositions \ref{thir:prop} and \ref{secon:propo}, then complexity in realizing time-advanced states is $\mathcal{O}(n \: \log n)$. 
\end{corollary}

\begin{proof}
As for any $n$, the product  ${\bf H}{\bf x}_k$ could be computed using the 2-FFTs \cite{SLARSN21,PERERA2025}, and hence the complexity in computing ${\bf H}{\bf x}_k$ to realize time advanced states cost $\mathcal{O}(n \: \log n)$. 
\end{proof}

\section{A Structured Neural Network (StNN) for Dynamical Systems}
\label{sec:StNN structure}
We show in this section that the Hankel operator can effectively predict time-advanced trajectories of dynamical systems using a low-complexity neural network, following the efficient learning and updating of the system’s dynamics. Thus, we introduce the StNN, showing its efficiency in training, learning, and updating dynamical systems, especially when compared with conventional feedforward neural networks. The StNN layers are designed using the matrix factorization of the Hankel operator (\ref{mformope}), which imposes significant constraints that minimize complexity and enhance performance. We begin with an overview of the StNN's construction, followed by its layer architecture, based on the matrix factorization of the Hankel operator (\ref{MLFFT}). Simply, we introduce an integration of model-based and data-driven learning with the design of StNN. 
The Figure \ref{fig:overview} illustrates the training and prediction process of the StNN for modeling the Lorenz system. The upper section represents the StNN Training phase, where Lorenz trajectories are used to generate a training dataset comprising input and output sequences. The StNN is trained to map past trajectory points to future states, learning the underlying dynamics of the system. The lower section depicts the StNN autoregressive predictions phase, where a trained StNN takes the initial condition of a random trajectory as input and iteratively predicts future states. This process results in a generated trajectory that closely follows the true Lorenz dynamics. The structured approach enhances the model's ability to capture chaotic behavior.

\subsection{Structured Neural Network Architecture}
We start the section with the forward propagation of the StNN, followed by its architecture. The forward propagation of the StNN leverages the revised factorization equation (\ref{MLFFT}), i.e., 
\begin{equation}
    \widehat{\bf H} \approx \tilde{\bf I}_{n}  [{\bf J}^T]_{n \times r} {F}_r \breve{\bf D}_{r} {F}_r [{\bf J}]_{r \times n},
    \label{rmeq}
\end{equation}
where symmetric sub-weight matrices $F_{r} \in \mathbb{R}^{r \times r}$ approximate ${\bf \mathfrak{F}}_r$, through customized layers that incorporate diagonal matrices and recursive FFT-like factorization. 
In other words, the layers of the network are designed to execute with real-valued inputs based on the factorization equation (\ref{MLFFT}) followed by a divide-and-conquer technique of $F_{r}$ as a real-valued replacement for  ${\bf \mathfrak{F}}_r$ through structured learning.
This approach is complemented by a layer-by-layer computation process utilizing matrix-vector products, which enables the StNN to achieve enhanced states effectively.

\begin{proposition}
\label{prop:strnn}
Let  \( {\bf x}_0 \in \mathbb{R}^{n \times 1} \) be the input vector, \( {\bf x}_{4} \in \mathbb{R}^{n \times 1} \) be the output vector, and $n$ be the number of states (nodes) in each layer
of a neural network. 
Let the output between the \( (i-1) \)-th and \( i \)-th hidden layer be given by:
\begin{equation}
    {\bf x}_{i} = \sigma_{i} (W_{i,i-1}{\bf x}_{i-1} + {\bf b}_{i})
    \label{eq:prop_equation}
\end{equation}
where  \( i :=\{1, 2, 3, 4\},  W_{i, i-1} \) is the weight matrix connecting the \((i-1)\)-th layer to the \( i \)-th layer, \( {\bf b} \) represents the bias vector, and \( \sigma \) is the activation function. Then, we can design a StNN to predict states \( {\bf x}_{k+1}  \) from \( {\bf x}_k \) using the weight matrices defined via $W_{1,0} \in \mathbb{R}^{pr \times n}, \quad W_{2,1} \in \mathbb{R}^{pn \times pr}, \quad W_{3,2} \in \mathbb{R}^{pn \times pn}, \quad W_{4,3} \in \mathbb{R}^{n \times pn},$ and their \( p \) number of parallel sub-weight matrices, denoted as \( w_{i,i-1} \), 
with the following structured weight matrices 

\begin{align*}
    &W_{1,0} = \begin{bmatrix}
         w_{1,0}\\
         w_{1,0}\\
        \vdots
        \\ 
        w_{1,0} \\
    \end{bmatrix}_{2pn \times n},
    W_{2,1} = \begin{bmatrix}
        w_{2,1} & 0 &... &0\\
        0 & w_{2,1} & ... &0 \\
        \vdots &\vdots & &\vdots \\
        0 &0 &... & w_{2,1}
    \end{bmatrix}_{pn \times 2pn},\\
    &W_{3,2} = \begin{bmatrix}
        w_{3,2} & 0 &... &0\\
        0 & w_{3,2} & ... &0 \\
        \vdots &\vdots & &\vdots \\
        0 &0 &... & w_{3,2}
    \end{bmatrix}_{pn \times pn}, \\
    &W_{4,3} = \begin{bmatrix}
         w_{4,3} & w_{4,3} &... &w_{4,3}
    \end{bmatrix}_{n \times pn},
\end{align*}
where $w_{1,0}={F}_{r} [{\bf J}]_{r \times n} 
\in \mathbb{R}^{r \times n}$, $w_{2,1} =[{\bf J}^T]_{n \times r}{F}_{r}\hat{D}_{r} \in \mathbb{R}^{n \times r}$,$ w_{3,2} =\tilde{\bf I}_{n} \in \mathbb{R}^{n \times n}$, $ w_{4,3} =  D_n \in \mathbb{R}^{n \times n}$,  ${F}_{r} = P_{r}^T \begin{bmatrix}
{F}_{n} & \\ 
 & {F}_{n}
\end{bmatrix}
H_{r}$ with random initialization of $2 \times 2$ weight matrices, 
$
P_{r}$ is an even-odd permutation matrix,
$
{H}_{r}=\left[\begin{array}{rr}
 {\bf I}_{n}  & {\bf I}_{n}\\ 
\grave{D}_{n}  &-\grave{D}_{n} \\ 
\end{array}\right]
$, $r=2n$, and $D_n$, $\hat{D}_n$ and $\grave{D}_n \in \mathbb{R}^{n \times n}$ are randomized diagonal weight matrices.   
\end{proposition}

\begin{proof}
Let us define the sub-matrices $w_{1,0} \in \mathbb{R}^{r \times n}$, $w_{2,1} \in \mathbb{R}^{n \times r}$,$ w_{3,2} \in \mathbb{R}^{n \times n}$, and $ w_{4,3} \in \mathbb{R}^{n \times n}$ based on the factorization of the Hankel operator (\ref{MLFFT})  followed by (\ref{rmeq}) and a divide-and-conquer technique to design layers and learn weights for the proposed network. We begin by grouping the matrices in the factorization (\ref{MLFFT}) followed by (\ref{rmeq}) into three distinct groups, followed by a random diagonal weight matrix, ensuring each group corresponds to the weight matrices connecting \((i-1)\)-th layer to the \( i \)-th layer. In the first hidden layer, we define $j$ parallel sub-weight matrices s.t. $w_{1,0} = F_r [{\bf J}]_{r \times n} 
$, 
to learn the weight matrix $W_{1,0}$. Next, the sub-weight matrices connecting the first and second hidden layers are defined by $w_{2,1} = [{\bf J}^T]_{n \times r} F_{r} \breve{\bf D}_{r}  $ which are utilized to learn the weight matrix $W_{2,1}$. Next, the sub-weight matrices between the second and third hidden layers are defined as $w_{3,2} = \tilde{\bf I}_{n}$, and we utilize those to learn the weight matrix $W_{3,2}$. 
After the third hidden layer, the sub-weight matrices connecting the last hidden layer to the output layer are represented as diagonal matrices, i.e., $w_{4,3} = D_n$. Consequently, a linear transformation based on diagonal weight matrices is applied to combine the outputs of the sub-weight matrices to learn the weight matrix $W_{4,3}$. In addition to these weight matrices, we have frozen all identity and zero matrices in the factorization of the Hankel operator (\ref{MLFFT}) followed by (\ref{rmeq}) at each network layer. This will enable us to develop a lightweight model. Also, we have not shared or reused matrices among different layers, ensuring that no additional matrices contribute to the network architecture. With this configuration of parallel sub-weight matrices and frozen matrices, along with the propagation described in equation (\ref{eq:prop_equation}), we efficiently train the weight matrices of the StNN using a lightweight model.  
\end{proof}



To illustrate the advantages of our proposed network architecture over the feed-forward neural network (FFNN), we will present the structure of the StNN alongside the FFNN followed by the flops count, as detailed in Table \ref{tab:layerwise_comparison_with_totals}.  

\subsection{Structured Neural Network Approach to Predict Trajectories of Dynamical Systems}
\label{sub:snnds}
To study the evolution of the dynamical system, we first focus on the simple Lotka-Volterra model, followed by the well-studied and highly chaotic Lorenz system. We compare StNN and LEADS for the Lotka-Volterra model and StNN, FFNN, SINDy, and HAVOK for the Lorenz system. Our goal is to compare their accuracy, flop counts, parameters, and long-term behavior to efficiently predict the time-advanced trajectories of the system.

The Lotka-Volterra model is defined via a set of nonlinear ODEs known as a "predator-prey" system and formulated as 
\begin{align*}
    \frac{dx}{dt} &= \alpha x - \beta x y, &
    \frac{dy}{dt} &= -\gamma y+ \delta xy,
\end{align*}
where $\alpha, \beta, \gamma, \delta$ are system parameters which define an \emph{environment} and $x$ and $y$ respectively represent prey and predator populations.

On the other hand, the Lorenz system is determined via a system of differential equations in the form

\begin{align}
    &\frac{dx}{dt} = \sigma(y-x),
    &\frac{dy}{dt} = x(\rho - z) - y,&
    &\frac{dz}{dt} = xy - \beta z
    \label{eq:lorenz_system}
\end{align}
where the state of the system is given by ${\bf x} = [x,y,z]^T$ with the parameters $\sigma = 10, \rho = 28, \text{ and } \beta = 8/3$. Thus, before starting the numerical simulations based on the StNN to solve the Lotka-Volterra model and predict time-advanced trajectories of the Lorenz system, we will cover the fundamentals of the proposed StNN.


To obtain the evolution of the Lorenz system, we generate a wide range of initial conditions, denoted by vector ${\bf x}_0$, and track the trajectories over time. We advance the initial conditions with a sampling time interval of $\Delta t$, which is not the actual time step. The next step is to acquire the matrices that represent the inputs and outputs of the system at states ${\bf x}_k$ and ${\bf x}_{k+1}$, respectively, with sample increments of $\Delta t$, which are correlated to ${\bf X}$ and ${\bf {X'}}$, respectively. These matrices are obtained by utilizing the trajectories that have been trained over time through the learned Hankel operator $\widehat{\bf H}$. Thus to capture the evolution of the non-linear nature of the dynamical systems, we use StNN and FFNN with 5 layers (when including the input, output, and 3 hidden layers) 
and different number of nodes in each layer while imposing the structure to the network using Propositions  \ref{secon:propo} and \ref{prop:strnn}, in order to carry on the forward propagation. The network will be trained on trajectories based on (\ref{bbeq}) to predict states in future time for any given initial conditions. The network will utilize activation functions such as Tanh and Sigmoid in the first two hidden layers and ReLU as the activation function of the third hidden layer to incorporate the dynamics of the system. As a result, we derive a new set of spatiotemporal data to generate future predictions from ${\bf x}_k$ to ${\bf x}_{k+1}$.

Additionally, we evaluate the training performance over $e$ epochs, using the loss function based on Proposition \ref{thir:prop} s.t.
\begin{equation}
    L({\bf x}_k,{\bf x}_{k+1}):= \frac{1}{m_b\times n}\| {\bf {X'}} - \widehat{\bf H} {\bf X}\|^2_F + \sum_{l=1}^{4} \alpha_l \|W_{l,l-1}\|_{\eta},
    \label{lossfunction}
\end{equation}
where $m_b$ represents the mini-batch size and $\alpha_l$ denotes the regularization hyperparameter that must be tuned for each layer
and hence validate the trajectory data of the trained model against the dynamical model using the best-fit time-advanced state-based Hankel operator $\widehat{\bf H}$.

\section{Numerical Simulations: Learn, Update, and Predict States}
\label{sec:experiments}
In this section, we first learn, update, and predict trajectories for the Lotka-Volterra model followed by the chaotic Lorenz system. Next, we compare numerical simulations based on the StNN and LEADS for the Lotka-Volterra model and StNN, FFNN, SINDy, and HAVOK for the Lorenz system.

\subsection{Numerical Simulations: Lotka-Volterra Model to Learn and Predict Dynamics}
\label{sec: lotka volterra}
\begin{figure*}[!t]
\centering

\subfloat[StNN Env. 1]{\includegraphics[width=0.48\textwidth]{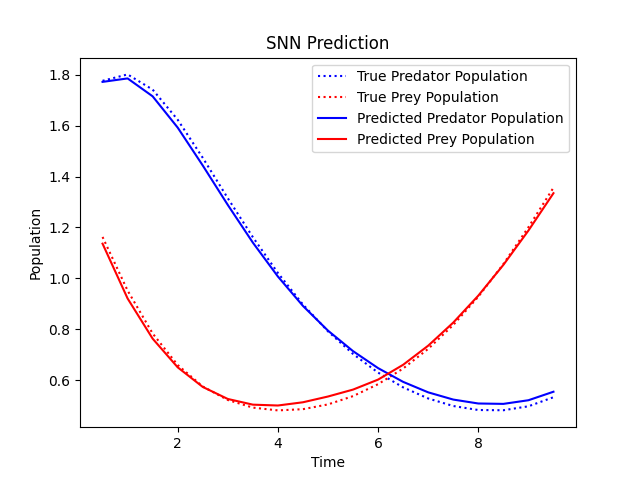}\label{fig:stnn_env1}}
\subfloat[LEADS Env. 1]{\includegraphics[width=0.48\textwidth]{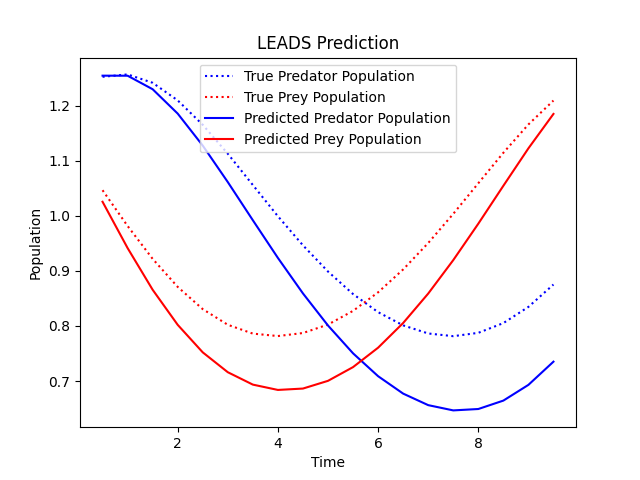}\label{fig:leads_env1}}
   
\subfloat[StNN Env. 10]{\includegraphics[width=0.48\textwidth]{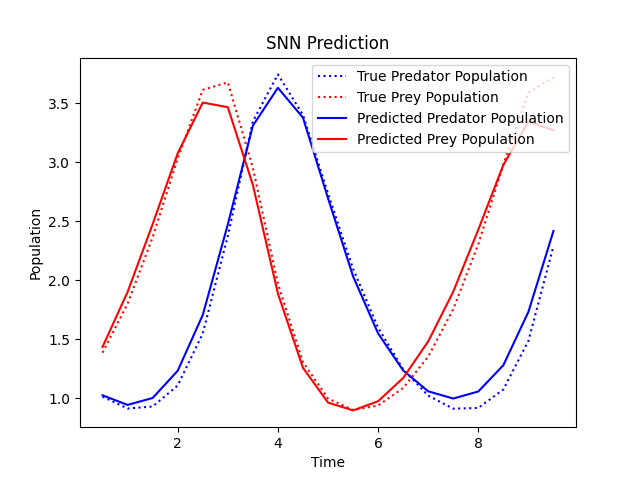}\label{fig:stnn_env10}}
\subfloat[LEADS Env. 10]{\includegraphics[width=0.48\textwidth]{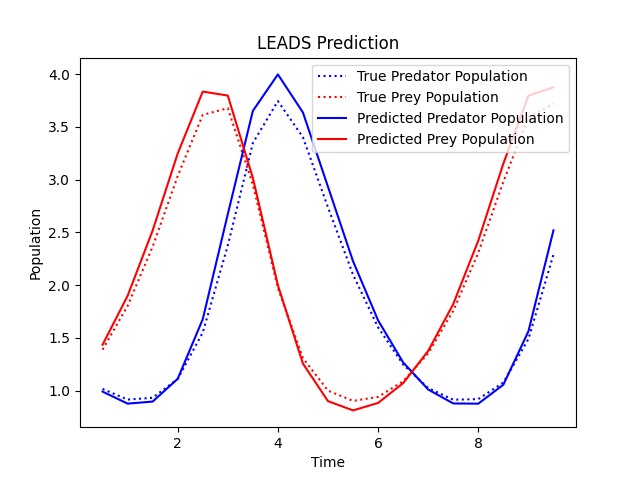}\label{fig:leads_env10}}

\caption{Autoregressive rollouts over 20-time steps of the respective models for the Lotka-Volterra system. While LEADS shows some divergence with the true solution at later times, StNN remains close to the true solution.}
\label{fig:SNN LEADS Lotka volterra}
\end{figure*}

In this section, we show numerical simulations to determine the time evolution of the Lotka-Volterra model for different \emph{environments}, where each environment is described by a set of system parameters $\alpha, \beta, \gamma$, and $\delta$.
In this experiment, we also draw comparisons with recently proposed model for dynamical systems, LEADS \cite{leads2021}. LEADS is a framework that leverages the commonalities and discrepancies among known environments to improve model generalization, using separate model components that focus either on global or environmental-specific dynamics. Following the experimental setup from LEADS, we consider 10 possible environments and generate trajectories each with 20 data points in time, $t_k = 0.0, 0.5, 1.0, \dots, 9.5$. For training, we sample 8 trajectories from each environment. Each environment has a unique set of system parameters, while each trajectory has a unique set of initial conditions. At evaluation, the models are tested on 32 trajectories from each environment. The model receives $x{(t_k)}$, $y(t_k)$, $t_k$ and the \emph{environment} passed as a unique integer which parametrizes the system parameters. The goal is to predict $x{(t_{k+1})}, y{(t_{k+1})}$ as outputs of the model. During the evaluation, the model only receives the initial conditions $x(t_0), y(t_0), t_0=0$, and environment specifier, performing an autoregressive rollout to predict all future $x, y$. The results of this experiment are summarized in Table \ref{tab:lotka volterra} and examples of predictions are provided in Figure \ref{fig:SNN LEADS Lotka volterra}.

\begin{table}[h!]
    \centering    
    \caption{Test results of the StNN on the Lotka-Volterra equations. Baseline experiments with the LEADS model \cite{leads2021} show that the proposed approach is able to obtain remarkable accuracy with very few parameters. }
    \begin{tabular}{|c|c|c|c|}
    \hline
        \textbf{Model} & \textbf{Training Time} & \textbf{Parameters} & \textbf{MSE}\\
        
    \hline
        StNN & 53 $s$ & 388 & $(2.02 \pm 0.39 )\times 10^{-3}$\\
        LEADS & 615 $s$ & 95095 & $(3.17 \pm 2.41)\times10^{-3}$ \\ \hline
    \end{tabular}
    
    \label{tab:lotka volterra}
\end{table}


   
    

Although LEADS was designed with novel elements to improve generalization across environments, we observe that large deviations from the truth may arise in some instances, illustrated in Figure \ref{fig:SNN LEADS Lotka volterra} (b). Meanwhile, StNN predicts dynamics which remain close to the ground truth. Additionally, Table \ref{tab:lotka volterra} illustrates several advantages of the StNN in parameter complexity and training time requirements. While LEADS has nearly 100,000 parameters, StNN is able to achieve a competitive error with a remarkable \emph{388 trainable parameters}. As a result, StNN may also be trained an entire order of magnitude faster than the competing approach. This experiment underlines the advantages of structured matrices with learnable parameters, as we propose in this work.

\subsection{Numerical Setup for the Chaotic Lorenz System}
\label{sec:Lorenzexperiments}
In this section, we analyze the performance of the StNN architecture compared to FFNN, SINDy, and HAVOK using the chaotic Lorenz system. 
To conduct these simulations, the Lorenz system, characterized by the differential equations (\ref{eq:lorenz_system}), was used to produce time-series data based on the parameters \(\sigma = 10\), \(\beta = \frac{8}{3}\), and \(\rho = 28\) with a time step of \(dt = 0.01\) at the duration of \(T = 8\), i.e. each trajectory consists of  800 data points. Furthermore, we obtained 100 such trajectories by perturbing the nominal initial state, i.e,  \([x(t_0), y(t_0), z(t_0)] = [0, 1, 20]\) with random uniform noise of magnitude 1. 
The \texttt{odeint} function from the SciPy library was used to numerically integrate each perturbed trajectory, guaranteeing high accuracy with relative and absolute tolerances set to \(1 \times 10^{-12}\). 
The StNN was implemented using a feedforward architecture with input, output, and three hidden layers, as explained in Section \ref{sub:snnds}. This model effectively combines activation functions (\(Tanh\), Leaky-ReLU, and \(ReLU\)) to capture the complex non-linear dynamics of the Lorenz system. The input and output dimensions were set to 4 by padding the state variables (\(x, y, z\)) with 0, i.e., (\(x, y, z, 0\))to match the dimensions.

An 80,000-sized randomly generated dataset was divided into input-output pairs, with each input being a state vector \([x_1(t_k), x_2(t_k), x_3(t_k), 0]\) and the output being the subsequent state vector \([x_1(t_{k+1}), x_2(t_{k+1}), x_3(t_{k+1}), 0]\). The input and output datasets were mini-batched $1000$ to ensure efficient mini-batch-wise training\cite{Goodfellow-et-al-2016}. We split the dataset into training and validation, i.e., 80\% of the dataset was allocated for training, while the remaining 20\% was reserved for validation. We utilized the Levenberg-Marquardt algorithm implemented in PyTorch by Di Marco \cite{dimarco2025torchlm}. This implementation enables efficient optimization for training neural networks by combining the advantages of gradient descent and Newton’s method. The training process was conducted over 20 epochs with 640 steps for each epoch, utilizing the high convergence rate of the Levenberg-Marquardt method for non-linear regression tasks. 
We set the hyperparameter value \(\alpha_2 = 1 \times 10^{-7}\) and \(\alpha_1, \alpha_3, \alpha_4 = 0\). The main reason for this selection is that we want to minimize the loss function (\ref{lossfunction}) with the order of \(10^{-6}\). Therefore, as the error approaches this order, we aim to balance the error term with the nuclear norm regularization term to enforce a low-rank structure in the StNN through the loss function \ref{lossfunction}.
To enhance readers' understanding of the theoretical foundation and its connection to the StNN learning algorithm, we direct readers to the code \href{https://github.com/Hansaka006/StNN-Dynamical-Systems}{StNN-Dynamical-Systems}.

 A summary of the training and validation performance for various StNN models with different \( p \) values is provided in Table \ref{tab:performance for different p}. The flops or parameter savings percentages are calculated in comparison to the FFNN using 
 
 \begin{align}
     &\text{flops (or parameter) saving} 
     := 
     \\ \nonumber
     &\frac{\#\text{FFNN(flops/parameters)} - \#\text{StNN(flops/parameters)}}{\#\text{FFNN(flops or parameters)}} \times 100\%.
     \end{align}
     \label{saveqn}

\begin{table}[h!]
\centering
\caption{We show the training and validation performance of StNN for different numbers of parallel \( p \) sub-weight matrix configurations based on \textbf{Table} \ref{tab:layerwise_comparison_with_totals}. This table summarizes the impact of varying \( p \) based on the loss function (\ref{lossfunction}) taken as training error, model weights, and computational complexity (flops). Savings percentages are calculated in the comparison of StNN to the FFNN using equation (\ref{saveqn}).}
\begin{tabular}{|c|c|c|c|c|}
\hline
\textbf{p} & \textbf{Training} & \textbf{Training}&\textbf{Parameter Saving} &\textbf{flops Saving}\\ & \textbf{Error (\ref{lossfunction})} &\textbf{Time} &\textbf{StNN vs FFNN} & \textbf{StNN vs FFNN }\\ 
& & (mins) &  &  \\ \hline
1 & $2.11\times10^{-2}$ & 4 & $96\%$ & $96\%$\\
2 & $6.40\times10^{-3}$ & 8  & $94\%$ & $92\%$\\ 
4 & $3.97\times10^{-4}$ & 13 & $87\%$ & $84\%$\\
6 & $1.96\times10^{-5}$ & 20  & $81\%$ & $76\%$\\ 
8 & $1.26\times10^{-6}$ & 27  & $74\%$ & $70\%$\\ \hline
\end{tabular}
\label{tab:performance for different p}
\end{table}


Smaller \( p \) values lead to higher loss, reflecting the trade-off between model simplicity and accuracy. While smaller \( p \) values reduce the number of weights and floating-point operations, they also result in less precise predictions over time. Conversely, larger \( p \) values yield significantly lower loss, highlighting their excellent predictive accuracy. However, this improvement comes at the cost of increased computational complexity, as seen in the greater number of weights and flops required. Interestingly, when compared to other StNN configurations, the best-performing StNN model is \( p = 6 \), which achieves a significant reduction in the error on the dataset. Furthermore, compared to the FFNN, this StNN achieves a reduction of roughly $81\%$ in the number of weights and $76\%$ in floating-point operations, demonstrating a significant advantage in computational and parameter complexity. 

Once the StNN and FFNN are trained and updated on the trajectory data, 
the non-linear dynamical model 
describing the Lorenz system could be to map the states from ${\bf x}_{k}$ to ${\bf x}_{k+1}$ and hence to predict the future states from an initial state. 

\begin{figure*}[ht]
    \centering
    \subfloat[FFNN autoregressive prediction over 5000-time steps]{\includegraphics[width=0.45\textwidth]{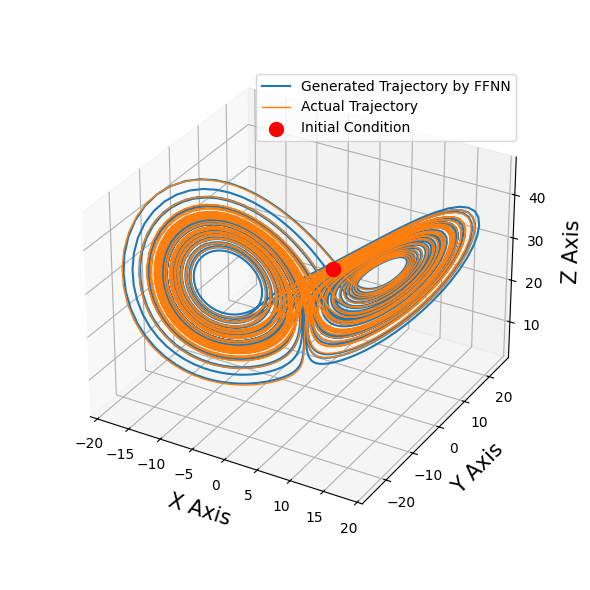}\label{fig:ann-3d}}
    \subfloat[StNN autoregressive prediction over 500-time steps]{\includegraphics[width=0.45\textwidth]{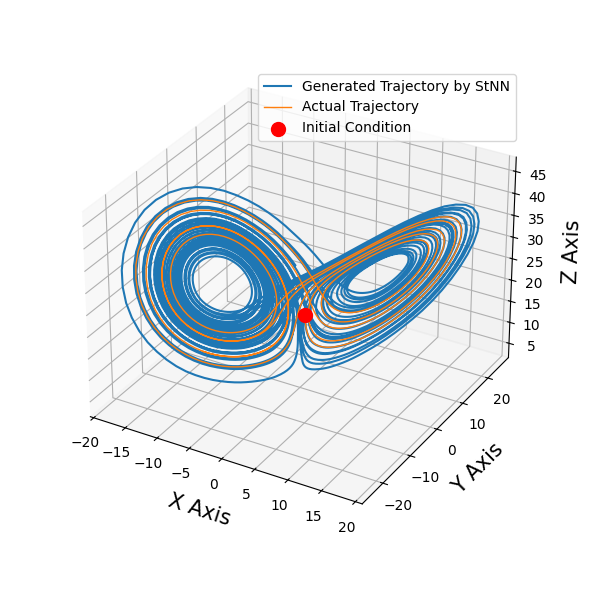}\label{fig:snn-3d}}
    \caption{Time-advanced trajectory prediction using trained FFNN and StNN models over 600-time steps. After training, the FFNN and StNN models are used to predict the future trajectory of the system given an initial condition (red marker). The left plot (a) shows the trajectory predicted by the FFNN (blue), while the right plot (b) shows the trajectory predicted by the StNN (blue). The actual trajectory (orange) serves as a reference for comparison. The results illustrate how well each model captures the system dynamics and maintains accuracy over extended time steps.}
    \label{fig:prediction_comparison_ANN_SNN}
\end{figure*}
Figure \ref{fig:prediction_comparison_ANN_SNN} was created using the trained StNN and FFNN to take an initial state and autoregressively advance the solution by $\Delta t$. The output at each time stamp was reinserted into the NNs to estimate the solution $k \Delta t$ to predict time-advanced states. This iterative mapping could produce a prediction for the future state as far into the future as desired. More specifically, figure \ref{fig:prediction_comparison_ANN_SNN} shows states mapping from  ${\bf x}_{k}$ to ${\bf x}_{k+1}$ to predict Lorenz solution 600-time steps into the future from a given initial state. The performance of the StNN was then compared with the FFNN to approximate the future dynamics of the system. The evolution of two randomly chosen trajectories is predicted using the StNN and FFNN as shown in figure \ref{fig:prediction_comparison_ANN_SNN}. Both networks show remarkable accuracy in predicting highly chaotic and non-linear dynamics to map states from  ${\bf x}_{k}$ to ${\bf x}_{k+1}$. To elaborate on this further, we also compare with SINDy and HAVOK, showing the time evolution of the individual components within the states ${\bf x}_{k}$ against the NNs prediction in the Section \ref{sub:comparison}. 
A short-term comparison table with StNN, FFNN, DMD, SiNDY, and HAVOK followed by a paragraph about it, should go here. 

\subsubsection{Comparisons and Predictions of StNN, FFNN, DMD, SINDy, and HAVOK}
\label{sub:comparison}
In this section, we utilize the StNN associated with the lowest validation loss, where \( p=6 \) to compare its performance against the FFNN, DMD, SINDy, and HAVOK models, focusing on accuracy and flop counts.  For this comparison, we use the benchmark simulations of the Lorenz system from DMD, SINDy, and HAVOK based on \cite{brunton2022data}.

\begin{table}[h!]
\centering
\caption{Comparison of training error, number of learnable parameters, and training time among StNN, FFNN, DMD, SINDy, and HAVOK.}
\begin{tabular}{|c|c|c|c|}
\hline
\textbf{Technique} & \textbf{Training Error \eqref{lossfunction}} & \textbf{Parameters} & \textbf{Training Time} \\
\hline
StNN & $1.26 \times 10^{-6}$ & $536$ & 27 min. \\
FFNN & $0.84 \times 10^{-6}$ & $2073$ & 72 min. \\
DMD & $2.48 \times 10^{2}$ & $15$ & 0.47 sec. \\
SINDy & $1.29 \times 10^{-6}$ & $7$ & 6 sec. \\
HAVOK & $8.59 \times 10^{-4}$ & $1825$ & 0.54 sec. \\
\hline
\end{tabular}
\label{tab:training_comparison}
\end{table}

Table \ref{tab:training_comparison} highlights the training error, number of learnable parameters, and training time among StNN, FFNN, DMD, SINDy, and HAVOK  of the Lorenz system. 
The proposed StNN reduces the number of parameters to just 536 and also the training time by more than 75\%, with an error order $10^{-6}$ in 27 minutes compared to FFNN with 2073 parameters and a longer training time of 72 minutes. Compared to FFNN and HAVOK, the proposed StNN is characterized by having the minimum number of parameters. Even though DMD requires the shortest training time compared to StNN, FFNN, SINDy, and HAVOK, it results in the highest error, showing that it does not accurately reflect the chaotic dynamics of the Lorenz system.
The HAVOK also trains quickly, but its higher training error compared to StNN reflects a trade-off between model simplicity and how well they fit the training data. Interestingly, SINDy achieves the lowest parameters, although it does not show prominent results for long-term behavior as of StNN shown in Table \ref{tab:inference_comparison}. 

\begin{figure*}[ht]
    \centering
    \subfloat[FFNN Model]{\includegraphics[width=0.45\textwidth]{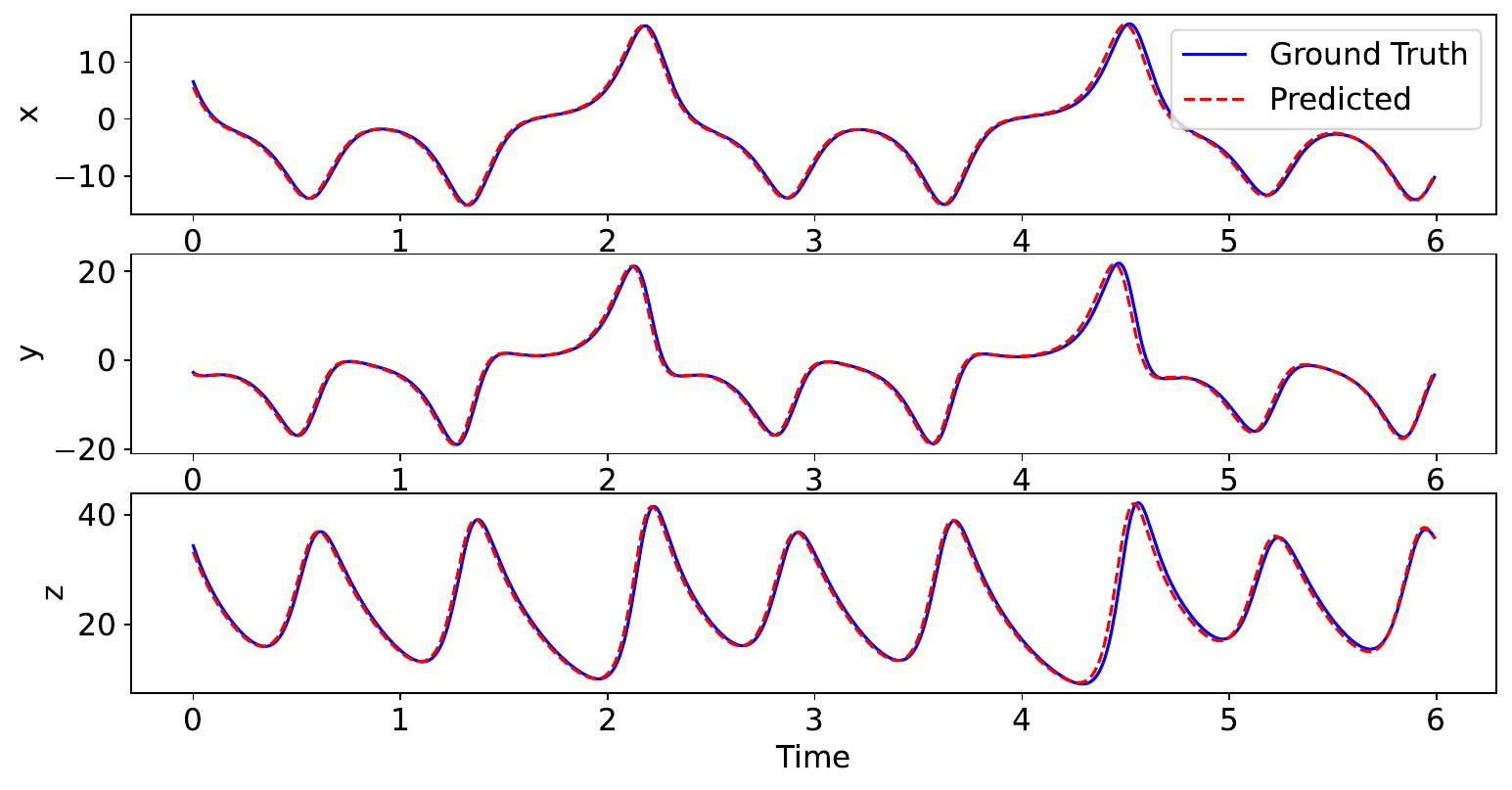}\label{fig:ann_time_advanced}}
    \subfloat[StNN Model]{\includegraphics[width=0.45\textwidth]{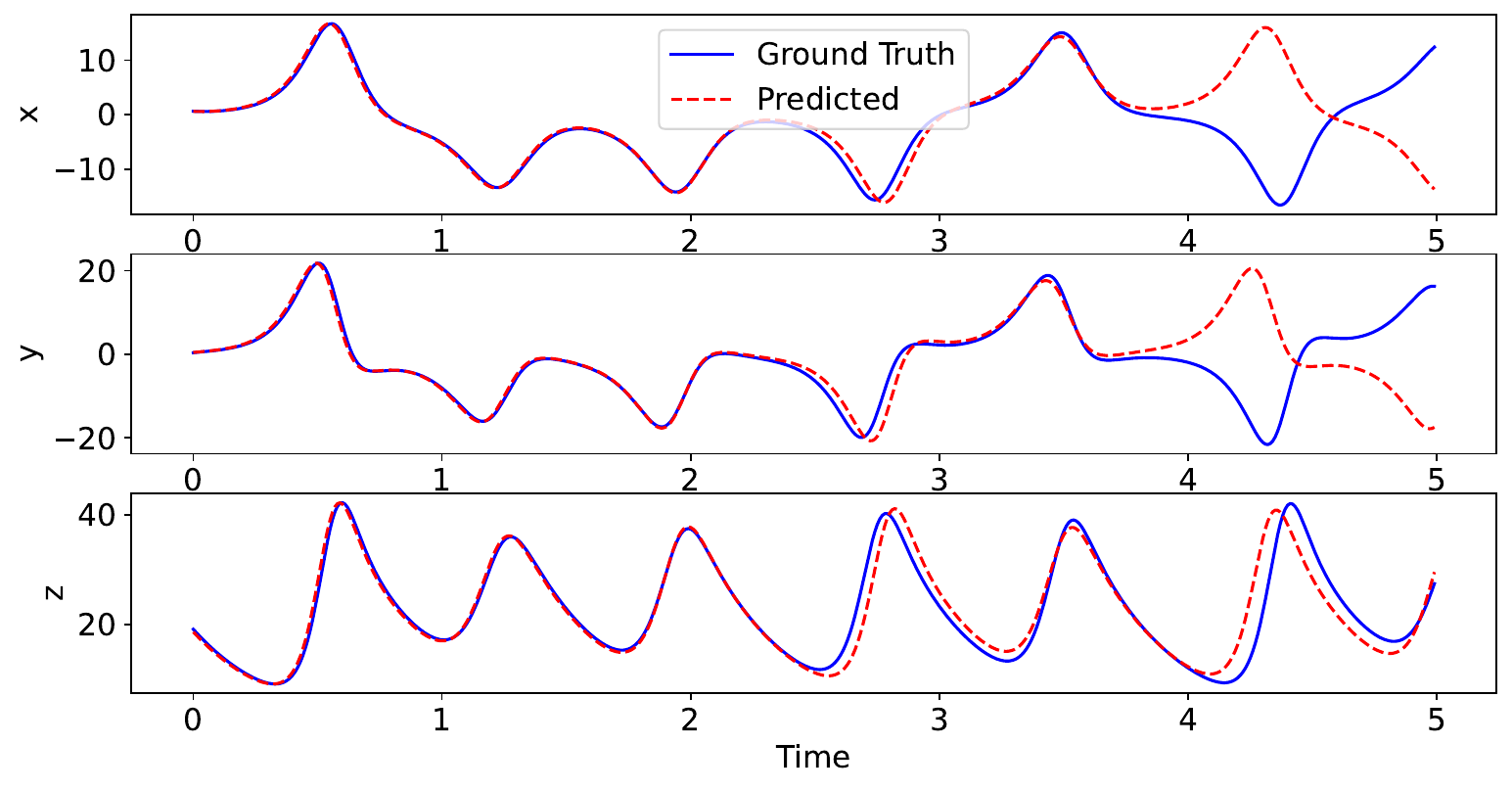}\label{fig:snn_time_advanced}}
    
    \subfloat[SINDy Model]{\includegraphics[width=0.45\textwidth]{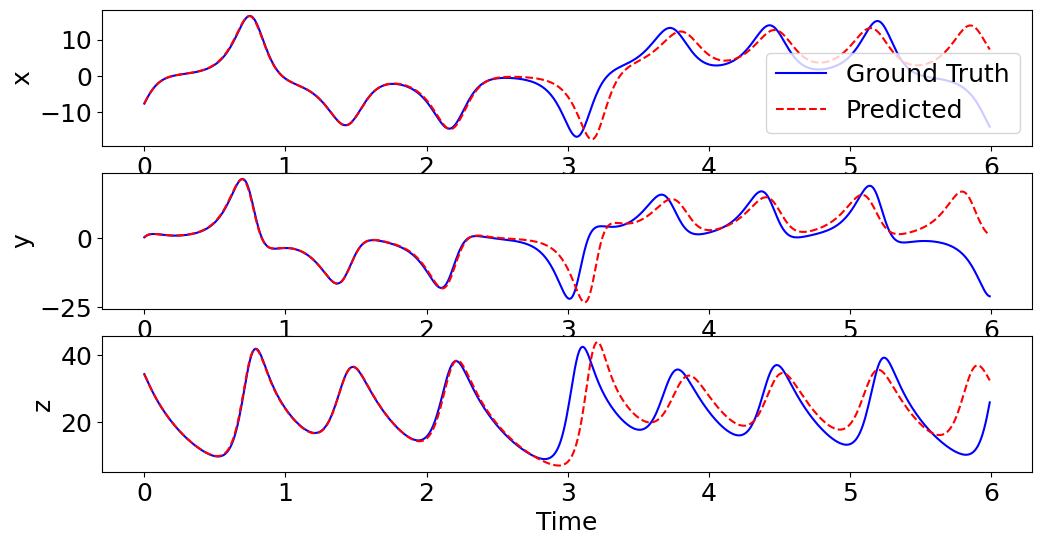}\label{fig:sindy}}
    \subfloat[HAVOK Model]{\includegraphics[width=0.45\textwidth]{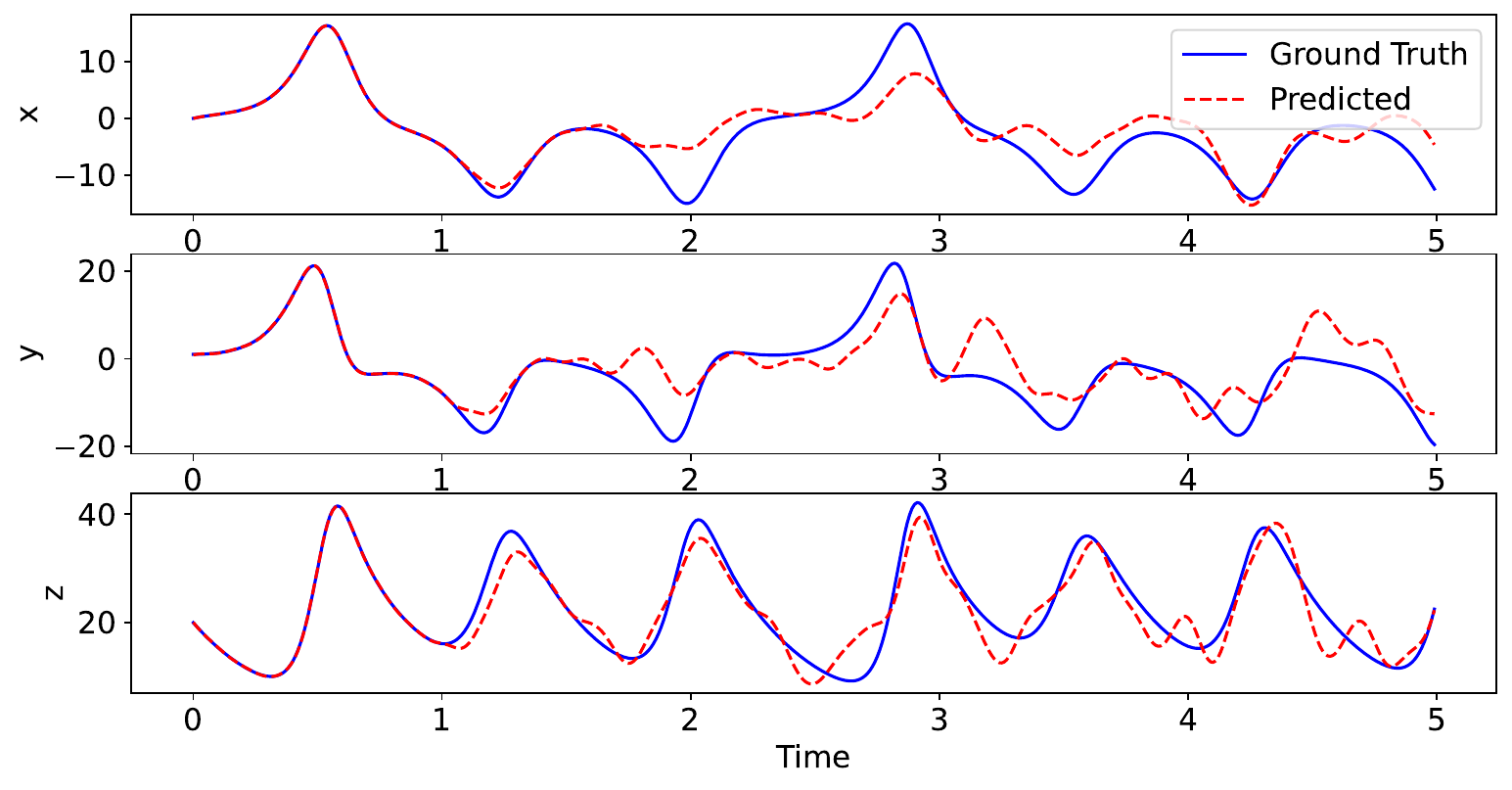}\label{fig:havok}}
    
    \subfloat[DMD Model]{\includegraphics[width=0.45\textwidth]{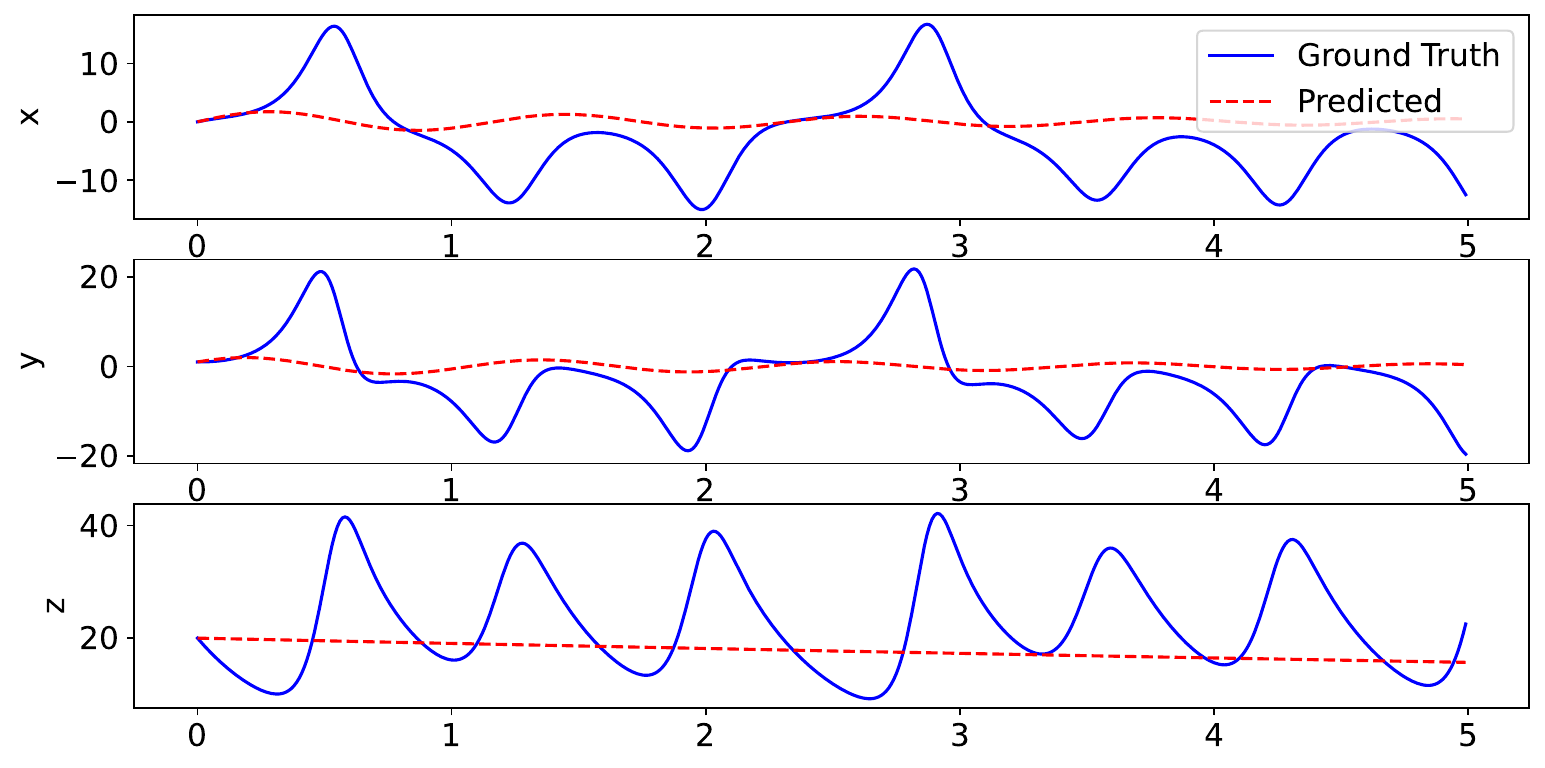}\label{fig:dmd}}
    \caption{Comparison of predicted and true trajectories along the x, y, and z axes over 500 time steps for different models. The FFNN (a) and StNN (b) models are the trained machine learning models used for trajectory prediction, while SINDy (c), HAVOK, (d) and DMD (e) serve as classical algorithm baselines for comparison with Python packages in \cite{desilva2020,demo2018pydmd, ichinaga2024pydmd}. The blue solid lines represent the true trajectory, while the red dashed lines indicate the predicted trajectory. The StNN followed by FFNN shows strong alignment with the true trajectory prediction, whereas DMD, SINDy, and HAVOK exhibit deviations. Among SINDy, DMD, and HAVOK, the DMD model fails to maintain stability, leading to extreme numerical divergence. This is since the DMD model is a simple linear operator by construction; it struggles to capture the nonlinear dynamics. In summary, this comparison shows that StNN is significantly more accurate in long-term predictions than conventional methods like DMD, SINDy, and HAVOK. 
    }
    \label{fig:MSE comparison of all}
\end{figure*}

For long-term predictions, we generated trajectories using the Lorenz equations with the same parameters discussed in Section \ref{sub:snnds} that were used to simulate the StNN and FFNN. The system is simulated up to  $T=50$ with a time step of \(dt = 0.01\), resulting in a dataset containing $5000$ time steps. Initial states \((x, y, z) = (0, 1, 20)\) were used to simulate the system, and the generated data was split into training and testing sets, with $80\%$ allocated for training and a subset of $500$ time steps 
for testing. We utilized the PySINDy Python package \cite{desilva2020} to simulate the SINDy model and the PyDMD Python package \cite{demo2018pydmd, ichinaga2024pydmd} to simulate the DMD and HAVOK models. Next, we created a 500-step random trajectory for the test, and we provided the FFNN, StNN, SYNDy, and HAVOK with the trajectory's initial condition. The 500-step trajectory was then iteratively predicted by running each model 500 times. The predicted values were compared with the actual values to calculate the MSE for 500 steps across three position values. 
Table \ref{tab:inference_comparison} and Figure \ref{fig:MSE comparison of all} show the time-advanced prediction of the Lorenz system based on StNN, FFNN, DMD, SINDy, and HAVOK.

\begin{table}[h!]
\centering
\caption{Predictions for the testing accuracy, inference flops, and inference time among StNN, FFNN, DMD, SINDy, and HAVOK. 
}
\begin{tabular}{|c|c|c|c|}
\hline
\textbf{Technique} & \textbf{Testing} & \textbf{Inference} & \textbf{Inference Time (ms)} \\
& \textbf{Accuracy \eqref{lossfunction}} & \textbf{flops} &\\
\hline
StNN & $2.43 \times 10^{-5}$ & 884 & 1.30 \\
FFNN & $0.99 \times 10^{-5}$ & $3960$ & 0.36 \\
DMD & $2.09 \times 10^2$ & $51$ & 0.19 \\
SINDy & $0.48 \times 10^{-5}$ & $209$ & 1.66 
\\
HAVOK & $1.90 \times 10^{-3}$ & $544$ & 0.20 \\
\hline
\end{tabular}
\label{tab:inference_comparison}
\end{table}

Based on Table \ref{tab:inference_comparison}, the proposed StNN requires fewer flops, i.e., 78\% reduced flops compared to FFNN with an accuracy order $10^{-5}$. Also, the accuracy in StNN is higher than that in HAVOK and much higher than in DMD. The inference time of StNN is lower than that of SINDy, but not as low as HAVOK and DMD. On the other hand, DMD has a higher error compared to the StNN and other models due to the linear approximation of the nonlinear model. 
Finally, we note here that FFNN, DMD, SINDy, and HAVOK benefit from PyTorch's highly efficient, GPU-accelerated operations. 
But, when we set $p=1$ in Table \ref{tab:performance for different p}, the proposed {\it StNN achieves an inference time of 0.238 ms} while standing out among FFNN, DMD, SINDy, and HAVOK. Also, based on the proposed StNN codes in \href{https://github.com/Hansaka006/StNN-Dynamical-Systems}{StNN-Dynamical-Systems}, StNN currently has not been implemented with low-level parallel optimization and has not been  benefited by PyTorch's libraries.  Certainly, the StNN codes can be parallelized, which will further minimize both flops and inference time compared to optimized PyTorch libraries in FFNN, DMD, SINDy, and HAVOK. Thus, compared to FFNN, DMD, SINDy, and HAVOK, StNN is a lightweight and low-complexity neural network suitable for modeling dynamical systems, making it particularly ideal for chaotic systems with well-defined governing equations.    
%


%

As shown in Fig. \ref{fig:MSE comparison of all}, for long-term predictions, SINDy accumulates a higher error compared to the StNN, particularly after 300 time steps. Thus, the SINDy tends to diverge from the predicted trajectory, while the StNN model continues to provide predictions that closely follow the trajectory for the given periods. Moreover, since the DMD model is a simple linear operator by construction, it struggles to capture the nonlinear dynamics of the Lorenz system. This results in a higher error of SINDy compared with StNN, as shown in Table \ref{tab:inference_comparison} and illustrated in Fig. \ref{fig:MSE comparison of all}.
Thus, the proposed StNN shows better performance while capturing complex, chaotic, and nonlinear dynamics of the Lorenz system compared to SINDy. Furthermore, StNN required a less computationally intensive training phase than FFNN with a smoother and lower cumulative loss compared to DMD, SINDy, and HAVOK, as shown in Fig. \ref{fig:MSE comparison of all}.
In terms of computational complexity, SINDy is lightweight, using fewer flops than StNNs. However, the StNN was shown to have lower inference time compared to SINDy and the best accuracy compared to SINDy, DMD, and HAVOK while precisely simulating the chaotic Lorenz system for long-term prediction.
In conclusion, the StNN outperforms all models in testing accuracy and inference time for long-term predictions of the chaotic Lorenz system. These qualities make the StNN an attractive choice for use in resource-constrained contexts where efficiency takes precedence as opposed to computationally expensive data-driven techniques. 

\section{Conclusions}
\label{sec:conclusions}
In this paper, we proposed a low-complexity structured neural network (StNN) for modeling and predicting the evolution of dynamical systems starting from the structured matrix theory. Our approach used the Hankel operator to give a structured and computationally efficient alternative to conventional neural networks and data-driven techniques like LEADs, DMD, SINDy, and HAVOK. According to numerical simulations based on the Lotka-Volterra model and the Lorenz system, the proposed StNN outperformed other methods in terms of decreasing complexity, training time, and inference time while retaining accurate long-term trajectory predictions. Our findings show that the structured nature of the Hankel operator-based neural network considerably decreases the number of parameters and flop counts while increasing the efficiency of the StNN when compared to conventional neural networks. Furthermore, comparisons to baseline approaches, such as FFNN, LEADs, DMD, SINDy, and HAVOK
demonstrate StNNs' promise for solving highly nonlinear and chaotic systems with highly accurate long-term predictions. 

Future work will include expanding the StNN framework to higher-dimensional dynamical systems and utilizing the Hankel operator defined through the observation of the state to efficiently solve PDEs.

\bibliographystyle{IEEEtran}
\bibliography{main}

\twocolumn[\newpage]
{\appendix[Layer-Wise Comparison of StNN and FFNN Models]
\begin{table}[htbp]
{\footnotesize
\centering
\caption{The StNN and FFNN architectures are designed for the layer-wise comparison of weight matrices, biases, total number of parameters, and flop counts. The value $p$ denotes the number of parallel sub-weight matrices designed for the values $p = 1, 2, 4, 6, 8$, which correspond to four distinct StNN models. These sub-matrices are used to construct and learn weight matrices $W_{i,i-1}$ that connect the \( (i-1) \)-th layer to the  \( i \)-th layer for $i=1,2,3,4$. }
\begin{tabular}{|c|c|c|c|c|c|c|c|} \hline
\textbf{Weight} & \textbf{Sub} & \textbf{Number of}& \textbf{Weights} & \textbf{Biases} & \textbf{Total} & \textbf{flop} \\ 
\textbf{Matrix} & \textbf{Weight} & \textbf{Parallel Sub}& & &\textbf{Parameters}& \textbf{Count} \\
& \textbf{Matrices}& \textbf{Weight Matrices}& & & &

\\\hline
\multicolumn{7}{|c|}{\textbf{StNN (Structured Neural Network)}} \\ \hline
$W_{1,0}$ & \begin{tabular}[c]{@{}c@{}}
    $[{F_{2}}]_{2 \times 2}$  \\
    $[{H]_{8 \times 8}}$ \\
    $[{H]_{4 \times 4}}$ 
\end{tabular} 
& \begin{tabular}[c]{@{}l@{}}
    $2p$ \\
    $p$ \\
    $p$ \end{tabular}
& \begin{tabular}[c]{@{}l@{}}
    $8p$ \\
    $4p$ \\
    $2p$ \end{tabular}
& $8p$ 
& $26p$ 
& $68p$ \\
\cline{2-7}
 & \text{Total} & - & $\mathbf{14p}$ & $\mathbf{8p}$ & $\mathbf{26p}$ & $\mathbf{64p}$ \\
\hline
$W_{2,1}$ & \begin{tabular}[c]{@{}c@{}}
    $[{\hat{D}}]_{8 \times 8}$ \\
    $[{F_{2}}]_{2 \times 2}$ \\
    $[{H]_{8 \times 8}}$ \\
    $[{H]_{4 \times 4}}$
\end{tabular} 
& \begin{tabular}[c]{@{}l@{}}
    $p$\\
    $2p$ \\
    $p$ \\
    $p$ \end{tabular}
& \begin{tabular}[c]{@{}l@{}}
    $8p$\\
    $8p$ \\
    $4p$ \\
    $2p$ \end{tabular}
& $4p$ 
& $26p$ 
& $68p$ \\
\cline{2-7}
 & \text{Total} & - & $\mathbf{22p}$ & $\mathbf{4p}$ & $\mathbf{26p}$ & $\mathbf{68p}$ \\
\hline
$W_{3,2}$ & \begin{tabular}[c]{@{}c@{}}
    $[\mathbf{\tilde{I}}]_{4 \times 4}$ \\
\end{tabular} 
& $p$
& \begin{tabular}[c]{@{}l@{}}
    $0$\\
    \end{tabular}
& $4p$ 
& $4p$ 
& $4p$ \\
\hline
$W_{4,3}$ & \begin{tabular}[c]{@{}c@{}}
    $[{{D}}]_{4 \times 4}$
\end{tabular}
& $p$
& \begin{tabular}[c]{@{}l@{}}
    $4p$
   \end{tabular}
& $4p$ 
& $8p$ 
& $12p-4$ \\
\hline
\textbf{Total} & - & - &$\mathbf{40p}$ & $\mathbf{20p}$ & $\mathbf{64p}$ & $\mathbf{148p-4}$ \\ 
\hline
\multicolumn{7}{|c|}{\textbf{FFNN (Feed-forward Neural Network)}} \\ \hline
$W_{1,0}$ & $[W]_{30\times3}$ & - & $90$ & $30$ & $120$ & $180$ \\
$W_{2,1}$ & $[W]_{30\times30}$ & - & $900$ & $30$ & $930$ & $1800$ \\
$W_{3,2}$ & $[W]_{30\times30}$ & - & $900$ & $30$ & $930$ & $1800$ \\
$W_{4,3}$ & $[W]_{3\times30}$ & - & $90$ & $3$ & $93$ & $180$ \\ \cline{1-7} 
\textbf{Total} & - & - & $1980$ & $93$ & $2073$ & $3960$ \\ \hline
\end{tabular}
\label{tab:layerwise_comparison_with_totals}
}
\end{table}
}
 





\end{document}